\pdfoutput=1
\documentclass[11pt]{article}

\usepackage[margin=1in]{geometry}
\usepackage{times}
\usepackage{microtype}
\usepackage{natbib}
\usepackage{hyperref}
\usepackage{url}
\usepackage{booktabs}
\usepackage{xcolor}
\usepackage{multirow}
\usepackage{graphicx}
\usepackage{wrapfig}

\usepackage{amsmath,amsfonts,bm}

\def\eqref#1{equation~\ref{#1}}

\def\1{\bm{1}}

\DeclareMathAlphabet{\mathsfit}{\encodingdefault}{\sfdefault}{m}{sl}
\SetMathAlphabet{\mathsfit}{bold}{\encodingdefault}{\sfdefault}{bx}{n}

\hypersetup{
    colorlinks=true,
    linkcolor=blue,
    citecolor=blue,
    urlcolor=blue,
    pdftitle={RoboMP-DINOv2: Prompts, Not Filters for Robust Robot Manipulation},
    pdfauthor={Han Qi and Heng Yang}
}

\title{RoboMP-DINOv2: Prompts, Not Filters for Robust Robot Manipulation}
\author{
Han Qi \qquad Heng Yang\\[0.35em]
Harvard School of Engineering and Applied Sciences\\
Harvard University
}
\date{}

\begin{document}
\maketitle

\begin{abstract}
Robot manipulation policies must generalize across visual distribution shifts
while retaining scene context that can affect the correct action.
General-purpose vision encoders are not designed specifically for visuomotor
control and may be sensitive to such shifts, whereas object-centric methods
often use segmentation masks as hard filters, discarding unmasked information
and making segmentation errors an information bottleneck. We propose \emph{RoboMP-DINOv2} (Robotics Mask-Prompted DINOv2), a
robot-oriented full-scene vision encoder that uses masks as spatial prompts rather
than visibility filters. RoboMP-DINOv2 extracts
dense DINOv2 features from the complete observation, injects learned
region-specific embeddings at mask-selected locations, jointly contextualizes
prompted and unprompted tokens, and aggregates them into a full-scene
representation for action prediction. To improve appearance generalization, we
combine RoboMP-DINOv2 with masked-region color randomization (MCR), denoting
the resulting training pipeline as \emph{RoboMP-DINOv2-MCR}. MCR randomizes
selected object regions during training while preserving contextual cues
elsewhere in the scene. Across seven simulated manipulation settings, RoboMP-DINOv2 achieves macro-average success rates of $60.7\%$ under spatial shifts and
$59.7\%$ under scene clutter, compared with $50.7\%$ and $41.0\%$,
respectively, for a DINOv2-based Diffusion Policy. Under appearance shifts
involving unseen object colors, RoboMP-DINOv2-MCR achieves $72.5\%$ average
success,
compared with $35.1\%$ for the strongest evaluated color-randomized baseline.
Experiments involving obstacle avoidance and decisions based on unmasked color
cues further demonstrate that the proposed designs improve robustness without
discarding behaviorally relevant scene information. Paired-observation
representation analyses additionally show lower sensitivity to the evaluated
clutter and appearance perturbations. Together, these results establish RoboMP-DINOv2 and
RoboMP-DINOv2-MCR as effective full-scene representation designs for robust robot
manipulation under the visual shifts considered in this work. Project code is
available at \url{https://github.com/han20192019/RoboMP_DINOv2}.
\end{abstract}

\section{Introduction}

Robot policies deployed beyond their training environments inevitably encounter
visual distribution shifts. The same manipulation problem may appear against a
different background, among unfamiliar objects, or with task objects of unseen
colors. A central challenge is that the magnitude of a visual change does not
necessarily reflect its behavioral relevance: a distractor may substantially
alter an observation without changing the desired action, whereas a small
obstacle entering the motion path may require a different trajectory. Effective
visual representations for manipulation must therefore balance task focus with
scene awareness, reducing sensitivity to nuisance variation while preserving
contextual information that may affect the correct behavior.

Modern robot policies commonly rely on visual encoders pretrained for
general-purpose image understanding. Models such as
DINOv2~\citep{oquab2023dinov2} provide strong transferable features, but their
pretraining objectives do not explicitly identify which image regions are
relevant to a particular manipulation problem or which visual changes should
leave the policy behavior unchanged. Object-centric methods introduce such
structure by using segmentation masks to isolate relevant entities in 2D images
or 3D point clouds~\citep{emukpere2025disentangled,zhu2023learning}. Although
this filtering can reduce sensitivity to irrelevant appearance variation, it
also turns segmentation into a hard information bottleneck: pixels or points
outside the selected regions are unavailable to the policy, even when they
describe an obstacle, a support surface, another robot arm, or other
behaviorally relevant context. Moreover, an inaccurate mask can directly
remove visual evidence from the representation. These limitations motivate a
different role for masks: rather than determining what the policy is allowed to
observe, masks can highlight selected regions while leaving the complete visual
scene accessible.

Following the principle of using masks as \emph{prompts, not filters}, we
propose \emph{RoboMP-DINOv2} (Robotics Mask-Prompted DINOv2), a robot-oriented
full-scene vision encoder that adapts DINOv2 to structured spatial guidance.
RoboMP-DINOv2 integrates four components: dense full-scene feature extraction, multi-region mask prompting, global
contextualization, and scene-level aggregation. It first extracts dense patch
features from the complete RGB observation using a pretrained DINOv2 backbone.
Learned region-specific embeddings are then added at locations identified by
the input masks, explicitly marking selected regions without removing or
replacing their underlying visual features. A lightweight transformer jointly
contextualizes prompted and unprompted tokens, allowing the representation to
capture interactions between selected objects and the surrounding scene.
Finally, the contextualized tokens are aggregated into a compact full-scene
representation for action prediction, and the complete encoder is fine-tuned
jointly with the policy. Unlike hard object-centric filtering, RoboMP-DINOv2 preserves
unmasked scene content and does not directly discard visual evidence when the
supplied masks are imperfect. The encoder itself consumes only an ordered
collection of binary masks and does not require object-category labels or
textual object descriptions as inputs.

Object appearance presents a complementary source of distribution shift. Even
when relevant regions are explicitly identified, a visual encoder may associate
an object with the colors observed during training. Conventional full-image
color randomization can weaken such correlations, but it may also modify color
cues that remain behaviorally relevant, such as an indicator light that
determines the correct action. The masks used by RoboMP-DINOv2 provide a natural
interface for localizing this augmentation. We therefore introduce
\emph{masked-region color randomization} (MCR), which randomizes RGB appearance
only within selected object regions while leaving the remainder of the scene
unchanged. MCR discourages reliance on training-specific object colors while
preserving scene geometry and unmasked contextual cues. Combined with the
full-scene representation produced by RoboMP-DINOv2, this localized
augmentation enables
the policy to generalize across task-object recoloring while remaining
sensitive to color information that is behaviorally relevant elsewhere in the
scene.

We evaluate RoboMP-DINOv2 with Diffusion Policy~\citep{chi2023diffusionpolicy}
across
seven simulated manipulation settings spanning object placement, obstacle
avoidance, peg insertion, tool use, long-horizon manipulation, and bimanual
stacking. Under held-out spatial configurations, RoboMP-DINOv2 achieves a
macro-average
success rate of $60.7\%$, compared with $50.7\%$ for a standard DINOv2-based
Diffusion Policy. Under scene clutter, the corresponding success rates are
$59.7\%$ and $41.0\%$. RoboMP-DINOv2 achieves higher observed success than
DP--DINOv2 across all seven settings in both regimes and compares favorably
with the evaluated 2D and 3D object-centric methods and other mask-based vision
encoders. We further evaluate RoboMP-DINOv2-MCR under appearance shifts involving
unseen
object colors. Among the evaluated encoder and color-augmentation combinations,
RoboMP-DINOv2-MCR matches the highest observed performance on one setting and
outperforms all alternatives on the remaining six, achieving a macro-average
success rate of $72.5\%$, compared with $35.1\%$ for the strongest evaluated
alternative. Experiments involving obstacle avoidance and decisions based on
unmasked color cues further demonstrate that the proposed designs preserve
scene information that remains relevant to control. Paired-observation analyses
also show that RoboMP-DINOv2 and RoboMP-DINOv2-MCR produce representations
with lower sensitivity
to the evaluated clutter and recoloring perturbations when the underlying
simulator state is held fixed. We interpret these findings as evidence for the
effectiveness of the complete RoboMP-DINOv2 and RoboMP-DINOv2-MCR designs
under the visual shifts
considered in this work.

Our contributions are twofold:

\begin{itemize}

    \item We propose \emph{RoboMP-DINOv2} (Robotics Mask-Prompted DINOv2), a
    robot-oriented full-scene vision encoder that integrates dense DINOv2
    features, multi-region mask prompting, global contextualization, and
    scene-level aggregation for visuomotor policy learning. RoboMP-DINOv2
highlights
    selected regions while preserving the complete observation, thereby
    avoiding the hard information bottleneck imposed by object-centric
    filtering.

    \item We introduce \emph{masked-region color randomization} (MCR) as a
    complementary training strategy. The resulting \emph{RoboMP-DINOv2-MCR} pipeline
    improves generalization to unseen object appearances while preserving
    behaviorally relevant contextual cues, and achieves the strongest observed
    overall performance among the evaluated encoder and augmentation
    combinations.

\end{itemize}
\section{Method}
\label{sec:method}

Our goal is to incorporate region-level spatial guidance into a pretrained
vision encoder without restricting the policy to a predefined set of segmented
regions. The proposed \emph{RoboMP-DINOv2} (Robotics Mask-Prompted DINOv2) encoder
comprises four stages: dense full-scene DINOv2 feature extraction, feature-space
mask prompting, global contextualization, and scene-level aggregation.
RoboMP-DINOv2 first extracts dense
visual features from the complete RGB observation and then injects learned
prompts at locations specified by the input masks. A lightweight transformer
jointly contextualizes all prompted and unprompted tokens before aggregating
them into a compact full-scene representation for the visuomotor policy. No
image regions or spatial tokens are removed at any stage of the encoder.

We additionally introduce masked-region color randomization in
Sec.~\ref{sec:color-randomization} as an optional training augmentation for
generalization to unseen object appearances. This augmentation complements
RoboMP-DINOv2 but is not part of the encoder architecture itself.

\subsection{Problem Formulation}
\label{sec:problem-formulation}

At time step $t$, the policy receives an RGB observation
$I_t \in \mathbb{R}^{3\times H_0\times W_0}$, robot proprioception $q_t$, a
language instruction $\ell$, and an ordered collection of $R$ binary masks,
\begin{equation}
    \mathcal{M}_t
    =
    \left(
        M_t^{(1)},\ldots,M_t^{(R)}
    \right),
    \qquad
    M_t^{(r)} \in \{0,1\}^{H_0\times W_0}.
    \label{eq:task-masks}
\end{equation}
The masks identify image regions associated with manipulated objects, goals,
tools, receptacles, or other selected entities. Their ordering is preserved:
each mask slot $r$ is associated with a learned prompt embedding, allowing the
encoder to learn a slot-specific interpretation from demonstrations. RoboMP-DINOv2
itself receives only the ordered binary masks; object-category labels and
textual object descriptions are not provided as encoder inputs.

The visual encoder $g_\psi$ maps the RGB observation and masks to a
$d$-dimensional full-scene representation,
\begin{equation}
    v_t
    =
    g_\psi(I_t,\mathcal{M}_t)
    \in \mathbb{R}^{d}.
    \label{eq:visual-encoder}
\end{equation}
The downstream policy $\pi_\theta$ is conditioned on $v_t$, proprioception, and
language. Importantly, the masks provide spatial guidance but do not remove
pixels, crop objects, or otherwise restrict which regions of the observation
remain available to the policy.

\paragraph{Mask acquisition.}
During training, task masks are obtained from simulator ground-truth
segmentation. At evaluation time, we do not assume access to simulator masks.
For each selected region, we use a text query with Grounding
DINO~\citep{liu2023groundingdino} to obtain candidate bounding boxes. After
non-maximum suppression, we retain the highest-confidence detection and use its
bounding box to prompt SAM~\citep{kirillov2023segment}. Among the masks returned
by SAM, we select the highest-confidence prediction. The text queries are used
only by this external mask-generation pipeline and are never provided to RoboMP-DINOv2
or the downstream policy. Before feature-space prompting, each predicted mask
is resized to the spatial resolution of the visual feature map using
nearest-neighbor interpolation. Thus, training uses simulator masks, whereas
evaluation relies on externally predicted masks.

\subsection{RoboMP-DINOv2: Mask-Prompted Full-Scene Encoding}
\label{sec:robomp}

\paragraph{Dense full-scene feature extraction.}
We resize $I_t$ to $H_I\times W_I$ and extract patch tokens using a Vision
Transformer backbone $\Phi$ initialized from
DINOv2~\citep{oquab2023dinov2}. Let $p$ denote the patch size,
$H_f=H_I/p$, $W_f=W_I/p$, and $N=H_fW_f$. The backbone produces $N$ patch
tokens, each with dimension $d_b$. We restore their two-dimensional spatial
arrangement and project them to the encoder dimension $d$ using a learned
$1\times1$ convolution $W_{\mathrm{proj}}$:
\begin{equation}
    F_t
    =
    W_{\mathrm{proj}}
    \left(
        \operatorname{reshape}
        \left(
            \Phi_{\mathrm{patch}}(I_t)
        \right)
    \right)
    \in
    \mathbb{R}^{d\times H_f\times W_f},
    \label{eq:dense-features}
\end{equation}
where $\Phi_{\mathrm{patch}}$ denotes the patch-token output of the visual
backbone.

Masks are not applied during feature extraction. Consequently, $F_t$ retains
visual information from the entire scene, including selected objects,
obstacles, support surfaces, the robot embodiment, distractors, and other
context outside the specified mask regions.

\paragraph{Feature-space mask prompting.}
For each mask slot $r$, we learn a prompt embedding
$e_r\in\mathbb{R}^{d}$. We resize $M_t^{(r)}$ to the feature resolution using
nearest-neighbor interpolation, obtaining
$\bar{M}_t^{(r)}\in\{0,1\}^{H_f\times W_f}$. Let
$P\in\mathbb{R}^{d\times H_f\times W_f}$ denote a learned two-dimensional
positional embedding. The prompted feature at spatial location $(u,v)$ is
\begin{equation}
    S_t[:,u,v]
    =
    F_t[:,u,v]
    +
    P[:,u,v]
    +
    \sum_{r=1}^{R}
    \bar{M}_t^{(r)}[u,v]\,e_r.
    \label{eq:mask-prompting}
\end{equation}

Equation~\ref{eq:mask-prompting} marks mask-selected locations by adding their
corresponding slot embeddings while retaining the underlying visual features.
Tokens outside all masks retain their visual and positional representations.
When multiple masks overlap, their prompt embeddings are added at the
overlapping locations.

This operation differs from hard object masking because no spatial token is
removed or replaced. It also differs from concatenating masks with the RGB
input because spatial guidance is introduced directly into the pretrained
feature space. The additive prompts should not be interpreted as fixed
attention weights. Instead, they provide learned region-specific conditioning
signals whose influence is determined by the subsequent contextualization
layers and the downstream policy objective.

\paragraph{Global contextualization.}
The appropriate action can depend on interactions between prompted regions and
unprompted scene context. For example, the trajectory for placing an object may
depend on whether an unmasked obstacle blocks the direct path. To model such
interactions, we flatten $S_t$ into $N$ spatial tokens and process them using a
lightweight transformer encoder $\mathcal{T}$ followed by layer normalization:
\begin{equation}
    Z_t
    =
    \operatorname{LN}
    \left(
        \mathcal{T}
        \left(
            \operatorname{flatten}(S_t)
        \right)
    \right)
    \in
    \mathbb{R}^{N\times d}.
    \label{eq:transformer}
\end{equation}
Self-attention allows information to propagate between prompted and unprompted
tokens, enabling the encoder to model relationships between selected entities
and the surrounding scene.

\paragraph{Full-scene aggregation.}
We aggregate the contextualized tokens using global average pooling:
\begin{equation}
    v_t
    =
    \frac{1}{N}
    \sum_{i=1}^{N} Z_{t,i}.
    \label{eq:scene-pooling}
\end{equation}
Unlike an object-only representation, $v_t$ summarizes the complete observation
after prompt-guided contextualization. It therefore exposes both the selected
regions and the surrounding scene to the downstream policy.

\paragraph{Architecture details.}
Our default RoboMP-DINOv2 implementation uses DINOv2 ViT-B/14, with backbone dimension
$d_b=768$ and patch size $p=14$. Input images are resized to
$112\times112$, producing an $8\times8$ feature grid. We set the encoder
dimension to $d=256$ and use three pre-normalization transformer encoder
layers, each with eight attention heads and a feed-forward dimension of $4d$.
The DINOv2 backbone is initialized from pretrained weights and jointly
fine-tuned with the projection layer, prompt embeddings, positional embedding,
contextualization transformer, output normalization, and downstream policy.

\subsection{Policy Conditioning and Optimization}
\label{sec:policy-training}

We encode the language instruction using a frozen CLIP text
encoder~\citep{radford2021clip}. Its normalized output is projected using a
learned multilayer perceptron $h_{\mathrm{lang}}$. At each time step, the
policy-conditioning feature is
\begin{equation}
    c_t
    =
    v_t
    \mathbin{\Vert}
    q_t
    \mathbin{\Vert}
    h_{\mathrm{lang}}
    \left(
        \operatorname{normalize}
        \left(
            \operatorname{CLIP}(\ell)
        \right)
    \right),
    \label{eq:policy-conditioning}
\end{equation}
where $\Vert$ denotes concatenation.

For an observation horizon $T_o$, we concatenate the most recent conditioning
features,
\begin{equation}
    C_t
    =
    c_{t-T_o+1}
    \mathbin{\Vert}
    \cdots
    \mathbin{\Vert}
    c_t,
    \label{eq:temporal-conditioning}
\end{equation}
and provide $C_t$ as global conditioning to Diffusion
Policy~\citep{chi2023diffusionpolicy}.

Let $A_t$ denote a demonstrated action sequence, $k$ a diffusion step, and
$\epsilon\sim\mathcal{N}(0,I)$ Gaussian noise. Following the standard forward
diffusion process, the noisy action sequence is
\begin{equation}
    A_t^{(k)}
    =
    \sqrt{\bar{\alpha}_k}\,A_t
    +
    \sqrt{1-\bar{\alpha}_k}\,\epsilon,
    \label{eq:forward-diffusion}
\end{equation}
where $\bar{\alpha}_k$ is determined by the diffusion noise schedule. The
noise-prediction network $\epsilon_\theta$ is optimized using
\begin{equation}
    \mathcal{L}_{\mathrm{policy}}
    =
    \mathbb{E}_{t,k,\epsilon}
    \left[
        \left\lVert
            \epsilon
            -
            \epsilon_\theta
            \left(
                A_t^{(k)},k,C_t
            \right)
        \right\rVert_2^2
    \right].
    \label{eq:policy-loss}
\end{equation}

We introduce no auxiliary feature-alignment or contrastive objective. RoboMP-DINOv2 and
the downstream Diffusion Policy are optimized jointly using only the policy
training objective in Eq.~\ref{eq:policy-loss}.

\subsection{Masked-Region Color Randomization}
\label{sec:color-randomization}

Feature-space prompting identifies selected regions but does not explicitly
prevent the encoder from exploiting correlations between their appearance and
the demonstrated actions. To improve generalization to unseen object colors, we
introduce \emph{masked-region color randomization} (MCR) as an optional training
augmentation for RoboMP-DINOv2.

Let
$\mathcal{R}_{\mathrm{aug}}\subseteq\{1,\ldots,R\}$ denote the mask slots
selected for randomization. We construct their union as
\begin{equation}
    M_t^{\mathrm{aug}}
    =
    \operatorname{clip}
    \left(
        \sum_{r\in\mathcal{R}_{\mathrm{aug}}}
        M_t^{(r)},
        0,
        1
    \right).
    \label{eq:augmentation-mask}
\end{equation}
For each training image, we sample a spatially constant RGB color
$a_t\sim\mathcal{U}([0,1]^3)$ and replace the pixels inside the union mask:
\begin{equation}
    \widetilde{I}_t
    =
    \left(1-M_t^{\mathrm{aug}}\right)\odot I_t
    +
    M_t^{\mathrm{aug}}\odot a_t,
    \label{eq:color-randomization}
\end{equation}
where masks and colors are broadcast over the channel and spatial dimensions as
needed. Equation~\ref{eq:color-randomization} is expressed in raw RGB space; in
implementation, we apply the corresponding transformation consistently with
the input normalization used by DINOv2.

The randomized image $\widetilde{I}_t$ is passed through the visual backbone,
while the original binary masks are retained for feature-space prompting in
Eq.~\ref{eq:mask-prompting}. In our experiments, a single sampled color is
applied to the union of the selected regions in each image. MCR therefore
changes the appearance of the selected regions while preserving their spatial
support, the scene geometry, and all unmasked visual context. MCR is applied
only during training; evaluation uses the original RGB observations.

We refer to the encoder trained without this augmentation as
\textbf{RoboMP-DINOv2}. When MCR is enabled during training, we denote the complete
encoder-and-augmentation pipeline as \textbf{RoboMP-DINOv2-MCR}. The two variants
share the same inference architecture; MCR changes only the training-time RGB
augmentation.
\section{Experiments}
\label{sec:experiments}

Our experiments evaluate the proposed approach under three forms of visual
distribution shift. We first evaluate \emph{RoboMP-DINOv2} under held-out spatial
configurations and added scene clutter. We then evaluate the
\emph{RoboMP-DINOv2-MCR} training pipeline under unseen task-object appearances. Finally, we compare RoboMP-DINOv2 with
closely matched architectural variants and analyze how strongly the learned
representations respond to the evaluated visual perturbations.

\subsection{Experimental Setup}

\begin{figure*}[t]
    \centering
    \includegraphics[width=\textwidth]{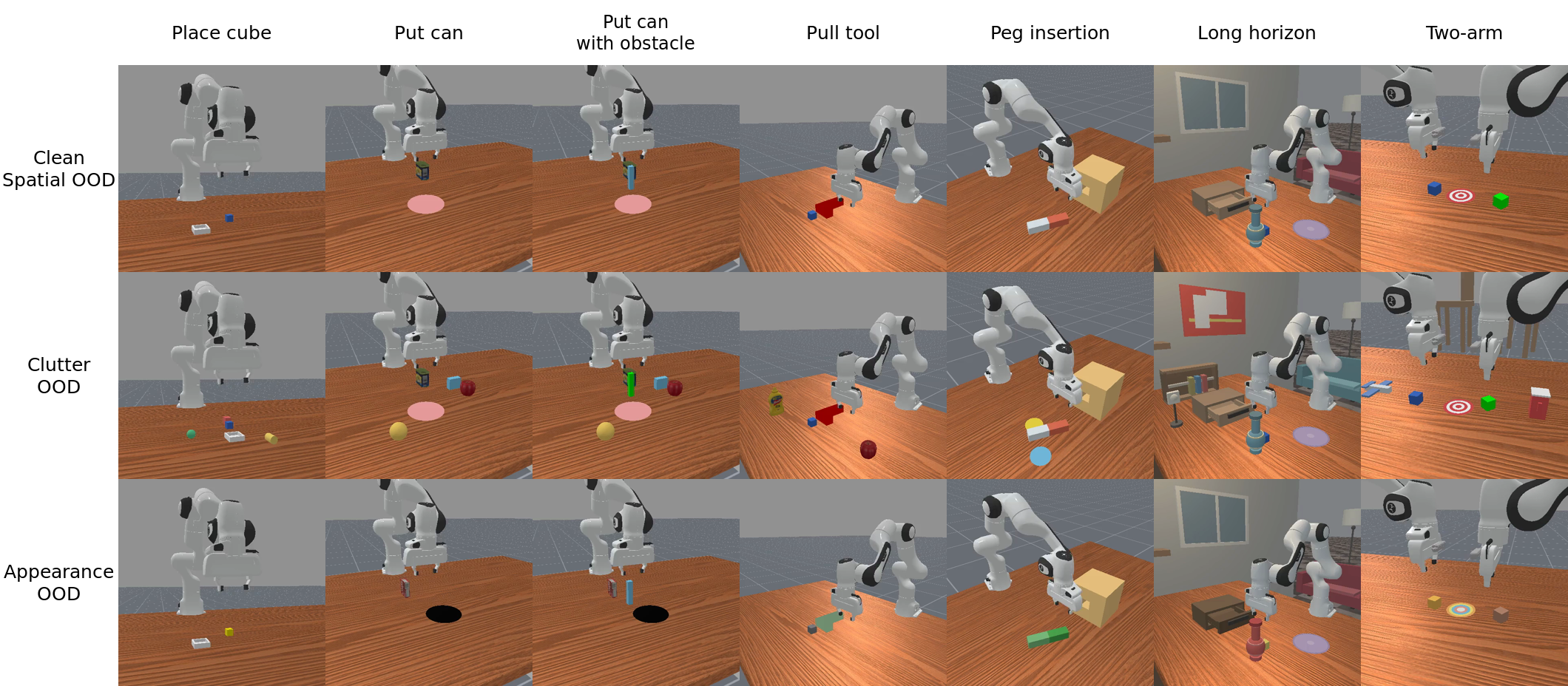}
    \caption{Seven simulated manipulation settings under the three evaluation
    regimes. Columns denote task settings, and rows show representative
    observations from Clean Spatial OOD, Clutter OOD, and Appearance OOD.}
    \label{fig:task_suite}
\end{figure*}

\paragraph{Task suite.}
We consider seven simulated manipulation settings
(Fig.~\ref{fig:task_suite}): Place Cube, Put Can, Put Can with an obstacle, Pull
Tool, Peg Insertion, Long Horizon, and Two-Arm cube stacking. Together, these
settings cover pick-and-place, obstacle avoidance, contact-rich insertion, tool
use, multi-stage interaction, and bimanual coordination.

The two Put Can variants are particularly useful for evaluating full-scene
representations. The obstacle-free and obstacle-present variants share the same
high-level goal but may require different motion trajectories. The policy
receives no explicit obstacle indicator and must infer from the visual
observation whether an obstacle blocks the direct path. A representation that
retains only the masked can and target regions may therefore omit information
needed to select the appropriate trajectory, whereas RoboMP-DINOv2 preserves the
surrounding visual context.

\paragraph{Evaluation regimes.}
All policies are trained using demonstrations collected in visually clean
scenes. We evaluate three out-of-distribution (OOD) regimes:

\begin{itemize}
    \item \textbf{Clean Spatial OOD:} Task objects and goals appear at held-out
    positions, while the surrounding scene retains the clean visual appearance
    observed during training.

    \item \textbf{Clutter OOD:} Additional task-irrelevant objects are inserted
    into the scene. Behaviorally relevant obstacles, when present as part of
    the task, remain relevant and must not be ignored.

    \item \textbf{Appearance OOD:} Task objects are rendered using colors not
    observed in the training demonstrations. For selected settings, objects
    are also initialized at held-out positions, combining appearance and
    spatial variation.
\end{itemize}

Clean Spatial OOD is already out of distribution with respect to object
configuration and should therefore not be interpreted as an in-distribution
reference condition.

\paragraph{Baselines.}
We compare RoboMP-DINOv2 with four representative visual-representation baselines.
\textbf{2D object-centric masking} follows the image-space decomposition of
DOCIR~\citep{emukpere2025disentangled}. It retains pixels corresponding to the
selected objects and robot arm while replacing the remaining image regions
with a uniform white background. \textbf{3D object-centric masking} follows the
object-level point-cloud representation of
GROOT~\citep{zhu2023learning}. It retains 3D points corresponding to the
selected objects and robot arm while filtering the remaining scene points.
\textbf{DP--DINOv2} combines a standard DINOv2 visual
encoder~\citep{hou2024diffusion} with Diffusion
Policy~\citep{chi2023diffusionpolicy}. \textbf{RoboGround--DINOv2} adapts the
grounded mask-conditioning mechanism of
RoboGround~\citep{huang2025roboground} to the DINOv2 policy setting by
concatenating the masks with the RGB observation. We denote our encoder by \textbf{RoboMP-DINOv2} and its training-time MCR
variant by \textbf{RoboMP-DINOv2-MCR}. For all DINOv2-based methods, the visual backbone
is jointly fine-tuned with the downstream policy.

\paragraph{Architecture and comparison protocol.}
All methods use the same 100 demonstrations per task, observation history,
proprioceptive and language inputs, downstream Diffusion Policy, optimization
objective, and training budget. The methods differ only in their visual
representations and method-specific preprocessing. For each baseline, we
preserve its default representation design rather than forcing all visual
encoders to use the same architecture or parameter budget.

All mask-based methods receive identical mask inputs: simulator ground-truth
masks during training and the same Grounding DINO+SAM
predictions~\citep{ren2024grounded} during evaluation. This protocol controls
for segmentation quality and focuses the comparison on how each representation
uses the available RGB and mask information.

\paragraph{Metrics.}
Our primary metric is task success rate, defined as the fraction of evaluation
rollouts satisfying the task-specific success criterion. We additionally
analyze the learned visual representations using paired clean and perturbed
observations that share the same underlying simulator state. Given a clean
feature vector $\mathbf{z}$ and its perturbed counterpart
$\widetilde{\mathbf{z}}$, we compute cosine distance and relative
$\ell_2$ distance:
\begin{equation}
    d_{\mathrm{cos}}
    =
    1-
    \frac{
        \mathbf{z}^{\top}\widetilde{\mathbf{z}}
    }{
        \lVert\mathbf{z}\rVert_2
        \lVert\widetilde{\mathbf{z}}\rVert_2
    },
    \qquad
    d_{\mathrm{rel}}
    =
    \frac{
        \lVert\mathbf{z}-\widetilde{\mathbf{z}}\rVert_2
    }{
        \lVert\mathbf{z}\rVert_2+\epsilon
    }.
    \label{eq:representation_distance}
\end{equation}
Both metrics normalize for feature magnitude and can be applied to
representations with different dimensionalities. Lower values indicate that
the evaluated visual perturbation produces a smaller representation change
when the underlying simulator state is fixed. These metrics diagnose
perturbation sensitivity; by themselves, they do not establish whether a
representation preserves all information required for control.

\paragraph{Scope of empirical claims.}
Our primary objective is to evaluate RoboMP-DINOv2 and RoboMP-DINOv2-MCR as complete
visual-representation designs rather than to estimate the causal effect of an
individual architectural or augmentation component. The main comparisons
therefore report the observed performance of complete encoder and preprocessing
pipelines. We additionally evaluate closely matched architecture and
augmentation variants to contextualize the proposed designs. We interpret
these comparisons as controlled empirical diagnostics rather than causal
estimates of the contribution of any single component.

\subsection{RoboMP-DINOv2 under Spatial and Clutter Shifts}
\label{sec:exp_mask_prompting}

We first evaluate RoboMP-DINOv2 without masked-region color randomization.

\subsubsection{System-Level Evaluation}

\paragraph{Evaluation scope.}
Table~\ref{tab:spatial_clutter_success} compares the complete RoboMP-DINOv2 architecture
with alternative visual representations under held-out spatial configurations
and added scene clutter. These experiments evaluate RoboMP-DINOv2 as a complete
representation design, including dense full-scene feature extraction,
multi-region mask prompting, global contextualization, and scene-level
aggregation.

\begin{table*}[t]
\centering
\caption{Success rates under held-out spatial configurations and scene clutter.
All methods are trained on demonstrations collected in visually clean scenes.
RoboMP-DINOv2 denotes the proposed mask-prompted full-scene encoder without
masked-region color randomization. ``Avg.'' is the macro-average over the seven
settings. The highest observed performance in each column and regime is shown
in bold.}
\label{tab:spatial_clutter_success}
\resizebox{\textwidth}{!}{
\begin{tabular}{@{}lcccccccc@{}}
\toprule
Method
& \shortstack{Place\\cube}
& \shortstack{Put can\\(clear)}
& \shortstack{Put can\\(obstacle)}
& \shortstack{Pull\\tool}
& \shortstack{Peg\\insertion}
& \shortstack{Long\\horizon}
& \shortstack{Two-\\arm}
& Avg. \\
\midrule
\multicolumn{9}{c}{\textit{Clean Spatial OOD}} \\
\midrule
2D object-centric masking
& .350 & .190 & .000 & .000 & .020 & .007 & .000 & .081 \\
3D object-centric masking
& \textbf{.850} & .840 & .070 & .000 & .085 & .036 & .120 & .286 \\
DP--DINOv2
& .410 & .820 & .750 & .150 & .170 & .812 & .440 & .507 \\
RoboGround--DINOv2
& .690 & .820 & .770 & .000 & .215 & .789 & .630 & .559 \\
RoboMP-DINOv2
& .540 & \textbf{.860} & \textbf{.830} & \textbf{.170}
& \textbf{.230} & \textbf{.848} & \textbf{.770} & \textbf{.607} \\
\midrule
\multicolumn{9}{c}{\textit{Clutter OOD}} \\
\midrule
2D object-centric masking
& .310 & .120 & .000 & .000 & .000 & .009 & .000 & .063 \\
3D object-centric masking
& \textbf{.780} & .260 & .030 & .000 & .000 & .018 & .060 & .164 \\
DP--DINOv2
& .340 & .590 & .520 & .130 & .130 & .751 & .410 & .410 \\
RoboGround--DINOv2
& .640 & .600 & .030 & .010 & .030 & .595 & .670 & .368 \\
RoboMP-DINOv2
& .560 & \textbf{.830} & \textbf{.800} & \textbf{.180}
& \textbf{.210} & \textbf{.787} & \textbf{.810} & \textbf{.597} \\
\bottomrule
\end{tabular}
}
\end{table*}

\paragraph{Clean Spatial OOD.}
RoboMP-DINOv2 achieves a seven-setting macro-average success rate of $60.7\%$, compared
with $55.9\%$ for RoboGround--DINOv2 and $50.7\%$ for DP--DINOv2. It achieves
the highest observed success on six of the seven settings, with the 3D
object-centric baseline performing best on Place Cube. These results
demonstrate the effectiveness of the complete RoboMP-DINOv2 architecture under held-out
spatial configurations.

The results also reveal a task-dependent pattern among the object-centric
baselines. The 3D object-centric baseline performs well on Place Cube, where
the information needed for control is concentrated within a small number of
selected objects. Its performance decreases substantially on settings
involving obstacle avoidance, tool use, insertion, long-horizon interaction,
and bimanual coordination. Similar limitations appear for the 2D
object-centric and RoboGround--DINOv2 baselines when information outside the
selected regions becomes important.

\paragraph{Clutter OOD.}
Under added scene clutter, RoboMP-DINOv2 achieves a macro-average success rate of
$59.7\%$, compared with $41.0\%$ for DP--DINOv2 and $36.8\%$ for
RoboGround--DINOv2. On obstacle-aware can placement, RoboMP-DINOv2 achieves $80.0\%$
success, compared with $52.0\%$ for DP--DINOv2 and $3.0\%$ for the 3D
object-centric baseline. In this setting, information outside the selected
object masks---particularly the obstacle and its relationship to the
surrounding scene---can directly affect the required trajectory. Hard
object-centric filtering can therefore remove information needed for control.

Manually detecting and masking every possible obstacle or distractor would
require a rigid, task-specific perception pipeline. RoboMP-DINOv2 instead retains the
complete scene and uses the supplied masks only as spatial prompts. Its
contextualization layers allow prompted objects to be represented jointly with
unprompted obstacles, distractors, and other scene content. RoboMP-DINOv2 also achieves
$81.0\%$ success on Two-Arm stacking, compared with $41.0\%$ for
DP--DINOv2.

\paragraph{Imperfect predicted masks.}
The qualitative examples in Fig.~\ref{fig:imperfect_masking} show that masks
predicted by Grounding DINO and SAM are not always perfectly aligned with the
selected objects. This reflects a practical setting in which evaluation-time
segmentation cannot be assumed to be exact. In hard object-centric
representations, mask errors can directly discard visual evidence by removing
regions outside the predicted masks. In RoboMP-DINOv2, mask errors alter the spatial
prompting signal but do not delete the underlying full-scene features.
Although these examples do not establish explicit robustness to controlled
mask corruption, they illustrate how RoboMP-DINOv2 avoids making predicted masks a hard
information bottleneck.

\subsubsection{Matched Architecture Comparisons}

\begin{figure}[t]
    \centering
    \includegraphics[width=\linewidth]{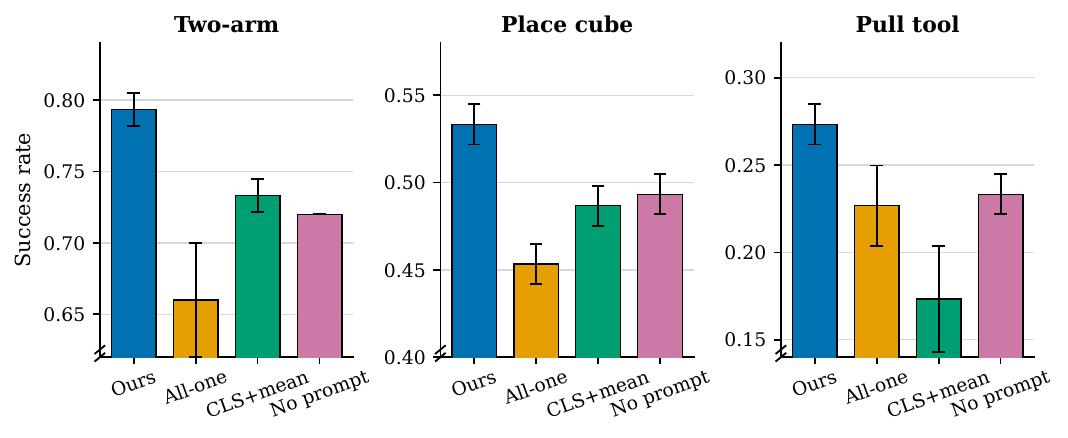}
    \caption{Matched architecture comparison under cluttered-scene evaluation.
    Bars show mean success over three training seeds evaluated on the same 50
    initial conditions; error bars show the standard deviation across seeds.
    All variants produce a 256-dimensional visual representation and use the
    same downstream policy. All-one masks retain prompting but remove spatial
    localization. No mask prompting removes the additive prompt term while
    retaining the remaining RoboMP-DINOv2 architecture. DINOv2 CLS+mean uses the global
    DINOv2 class token and mean-pooled patch features without the
    contextualization transformer.}
    \label{fig:ablation_architecture}
\end{figure}

The preceding experiments compare complete encoder designs. To examine RoboMP-DINOv2
relative to closely related architectural alternatives, we compare it with
three controlled variants on representative tasks, as shown in
Fig.~\ref{fig:ablation_architecture}:

\begin{itemize}
    \item \textbf{All-one masks} replaces every segmentation mask with an
    all-one mask, causing the learned prompt embeddings to be added uniformly
    to all patch tokens. This variant retains the prompt parameters and
    full-scene contextualization transformer while removing the spatial
    localization supplied by the masks.

    \item \textbf{No mask prompting} retains the DINOv2 backbone, projection
    layer, positional embedding, three-layer transformer, normalization, and
    global pooling used by RoboMP-DINOv2 but removes the additive prompt term in
    Eq.~\ref{eq:mask-prompting}.

    \item \textbf{DINOv2 CLS+mean} concatenates the DINOv2 class token with the
    mean-pooled patch token and maps the resulting 1536-dimensional feature to
    256 dimensions using a linear layer, LayerNorm, and GELU.
\end{itemize}

\paragraph{Results.}
RoboMP-DINOv2 achieves the highest observed mean success across all three evaluated
settings. The all-one-mask and no-prompt variants provide closely matched
reference points in which spatial localization or additive prompting is
removed while most of the remaining encoder architecture is retained. The
DINOv2 CLS+mean variant provides a conventional global-readout reference.
Together, these comparisons support RoboMP-DINOv2 as a complete encoder architecture in
the evaluated settings. We interpret the differences as comparative evidence
about the complete architectural variants rather than as causal estimates of
any individual component.

\subsubsection{Feature Consistency under Clutter}

We next examine how strongly added clutter changes the learned representation
when the underlying simulator state is fixed.
Table~\ref{tab:feature_distance} reports the distances between paired clean and
perturbed features. RoboMP-DINOv2 produces lower cosine and relative $\ell_2$ distances
than DP--DINOv2 on four of the five evaluated settings. The exception is
Two-Arm stacking, for which DP--DINOv2 produces slightly lower distances under
clutter.

These results show that, under the evaluated training configurations, RoboMP-DINOv2 is
generally less sensitive than DP--DINOv2 to the addition of scene clutter.
This analysis characterizes the representations produced by the complete
encoders; it does not isolate the contribution of an individual RoboMP-DINOv2
component.

\begin{table*}[t]
\centering
\caption{Feature distances between perturbed observations and their paired
clean counterparts (mean $\pm$ standard deviation over $N$ observation pairs).
Cosine distance is reported in units of $10^{-3}$ and relative $\ell_2$
distance in units of $10^{-2}$. DP--DINOv2 uses a 768-dimensional feature,
whereas RoboMP-DINOv2 uses a 256-dimensional feature. Under Clutter, RoboMP-DINOv2 denotes the
encoder trained without color randomization. Under Appearance, DP--DINOv2 is
trained without color randomization, whereas RoboMP-DINOv2-MCR denotes the complete
encoder-and-augmentation pipeline. Lower values indicate smaller
representation changes under the evaluated perturbation; the lower mean is
shown in bold.}
\label{tab:feature_distance}
\scriptsize
\setlength{\tabcolsep}{5pt}
\renewcommand{\arraystretch}{1.08}
\begin{tabular}{llrcccc}
\toprule
& & & \multicolumn{2}{c}{Cosine $\downarrow$}
& \multicolumn{2}{c}{Relative $\ell_2$ $\downarrow$} \\
\cmidrule(lr){4-5}\cmidrule(lr){6-7}
Task & Shift & $N$ & DP--DINOv2 & \shortstack{RoboMP-DINOv2\\/ -MCR}
& DP--DINOv2 & \shortstack{RoboMP-DINOv2\\/ -MCR} \\
\midrule
Out-of-bin
& Clutter
& 1269
& $31.770\!\pm\!29.030$
& $\mathbf{2.880\!\pm\!2.990}$
& $22.938\!\pm\!10.619$
& $\mathbf{6.707\!\pm\!3.555}$ \\
& Appearance
& 1269
& $21.752\!\pm\!30.249$
& $\mathbf{.0028\!\pm\!.0064}$
& $16.784\!\pm\!12.520$
& $\mathbf{.181\!\pm\!.150}$ \\
\midrule
Pull tool
& Clutter
& 2409
& $.002\!\pm\!.005$
& $\mathbf{.001\!\pm\!.002}$
& $.153\!\pm\!.109$
& $\mathbf{.126\!\pm\!.077}$ \\
& Appearance
& 2409
& $.240\!\pm\!.485$
& $\mathbf{.007\!\pm\!.037}$
& $1.856\!\pm\!1.509$
& $\mathbf{.230\!\pm\!.281}$ \\
\midrule
Peg insertion
& Clutter
& 1514
& $.625\!\pm\!1.347$
& $\mathbf{.249\!\pm\!.546}$
& $2.642\!\pm\!2.373$
& $\mathbf{1.636\!\pm\!1.516}$ \\
& Appearance
& 1514
& $.553\!\pm\!1.621$
& $\mathbf{.013\!\pm\!.042}$
& $2.073\!\pm\!2.634$
& $\mathbf{.375\!\pm\!.358}$ \\
\midrule
Long horizon
& Clutter
& 4740
& $.105\!\pm\!.115$
& $\mathbf{.040\!\pm\!.091}$
& $1.296\!\pm\!.700$
& $\mathbf{.665\!\pm\!.601}$ \\
& Appearance
& 4740
& $.216\!\pm\!.313$
& $\mathbf{.002\!\pm\!.005}$
& $1.658\!\pm\!1.322$
& $\mathbf{.170\!\pm\!.138}$ \\
\midrule
Two-arm
& Clutter
& 3028
& $\mathbf{1.769\!\pm\!1.315}$
& $2.025\!\pm\!1.576$
& $\mathbf{5.595\!\pm\!2.141}$
& $5.874\!\pm\!2.449$ \\
& Appearance
& 3028
& $13.588\!\pm\!27.961$
& $\mathbf{.019\!\pm\!.072}$
& $11.813\!\pm\!15.161$
& $\mathbf{.371\!\pm\!.494}$ \\
\bottomrule
\end{tabular}
\end{table*}

\subsection{Appearance Generalization with Masked-Region Color Randomization}
\label{sec:exp_appearance}

We next evaluate robustness to unseen task-object colors. We first use the
Appearance rows of Table~\ref{tab:feature_distance} to measure how strongly
recoloring the task objects changes the learned representation when the
underlying simulator state is fixed. DP--DINOv2 is trained without color
randomization in this representation-level comparison. Its features change
substantially under task-object recoloring, even though the scene geometry and
required action remain unchanged. In contrast, the complete RoboMP-DINOv2-MCR pipeline
reduces relative $\ell_2$ distance by $92.9\times$ on Out-of-bin,
$8.1\times$ on Pull Tool, $5.5\times$ on Peg Insertion, $9.8\times$ on Long
Horizon, and $31.8\times$ on Two-Arm stacking. The same overall trend appears
in cosine distance.

This comparison shows that, under the evaluated training configurations,
task-object recoloring produces substantially smaller representation changes
for RoboMP-DINOv2-MCR than for unaugmented DP--DINOv2. Because the two pipelines differ
in both encoder architecture and training augmentation, the comparison
characterizes their resulting perturbation sensitivity without attributing the
difference to either component individually.

Table~\ref{tab:appearance_ood} provides a system-level behavioral comparison in
which every method is trained with color randomization. Because the standard
DP--DINOv2 baseline does not use task masks, its randomization is applied to
the complete image. For mask-based methods, randomization is restricted to the
selected mask regions. RoboMP-DINOv2-MCR achieves the highest observed success in all
seven settings, tying RoboGround--DINOv2 on Place Cube and outperforming it on
the remaining six. Its seven-setting macro-average is $72.5\%$, compared with
$35.1\%$ for RoboGround--DINOv2 and $19.1\%$ for DP--DINOv2. Under this
comparison protocol, RoboMP-DINOv2-MCR achieves the strongest overall performance among
the evaluated complete pipelines.


\begin{table*}[t]
\centering
\caption{Success rates under Appearance OOD. All methods are trained with color
randomization. DP--DINOv2 uses full-image randomization, whereas the mask-based
methods apply randomization only within selected mask regions. ``Avg.'' is the
macro-average over the seven settings. The highest observed performance in
each column is shown in bold.}
\label{tab:appearance_ood}
\resizebox{\textwidth}{!}{
\begin{tabular}{lcccccccc}
\toprule
Method
& \shortstack{Place\\cube}
& \shortstack{Put can\\(clear)}
& \shortstack{Put can\\(obstacle)}
& \shortstack{Pull\\tool}
& \shortstack{Peg\\insertion}
& \shortstack{Long\\horizon}
& \shortstack{Two-\\arm}
& Avg. \\
\midrule
2D object-centric masking (masked-area rand.)
& .790 & .040 & .010 & .000 & .000 & .003 & .000 & .120 \\
3D object-centric masking (masked-area rand.)
& .940 & .030 & .000 & .000 & .010 & .015 & .000 & .142 \\
DP--DINOv2 (full-image rand.)
& .330 & .050 & .100 & .100 & .250 & .498 & .010 & .191 \\
RoboGround--DINOv2 (masked-area rand.)
& \textbf{.990} & .650 & .660 & .000 & .000 & .160 & .000 & .351 \\
RoboMP-DINOv2-MCR
& \textbf{.990} & \textbf{.710} & \textbf{.670} & \textbf{.820}
& \textbf{.700} & \textbf{.906} & \textbf{.280} & \textbf{.725} \\
\bottomrule
\end{tabular}
}
\end{table*}

\paragraph{Robustness under simultaneous appearance and clutter shifts.}
We next consider a more challenging condition in which appearance and clutter
shifts occur simultaneously. Table~\ref{tab:appearance_ablation} evaluates
Two-Arm stacking with both unseen object colors and additional scene clutter.
DP--DINOv2 with full-image randomization obtains a mean score of $0.01$.
Restricting the augmentation to selected mask regions, without providing the
masks to the encoder, improves the score to $0.11$. This result indicates that
the spatial support of the augmentation is consequential even in the absence
of mask-conditioned encoding. Combining masked-region randomization with
RoboGround--DINOv2 and 2D object-centric masking produces scores of $0.07$ and
$0.00$, respectively. RoboMP-DINOv2-MCR achieves the highest score, $0.20$, among the
evaluated combinations.


\begin{table}[t]
\centering
\small
\setlength{\tabcolsep}{4pt}
\caption{Appearance-randomization comparison under combined appearance-and-clutter shifts. Values report mean success rates across the three representative tasks.}
\label{tab:appearance_ablation}
\begin{tabular}{@{}lccc@{}}
\toprule
Method & Two arm & Place cube & Pull tool \\
\midrule
DP + full-image rand. & 0.00 & 0.02 & 0.00 \\
DP + masked rand. & 0.14 & 0.08 & 0.34 \\
Only mask + mask rand. & 0.00 & 0.04 & 0.00 \\
RoboMP-DINOv2-MCR & 0.26 & 0.24 & 0.38 \\
\bottomrule
\end{tabular}
\end{table}

\begin{figure}[htp]
    \centering
    \includegraphics[width=\linewidth]{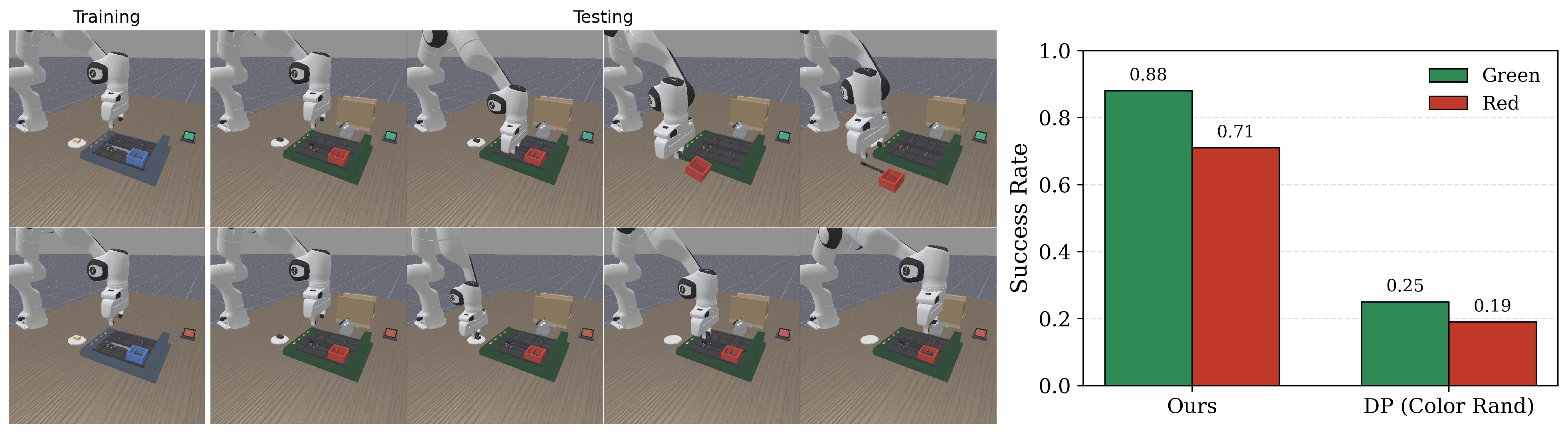}
    \caption{Preserving behaviorally relevant color cues outside the randomized
    regions. The stove-light color determines the correct action. RoboMP-DINOv2-MCR
    retains access to this contextual signal and substantially outperforms
    DP--DINOv2 trained with full-image color randomization under both light
    conditions.}
    \label{fig:kitchen_color_cue}
\end{figure}

\paragraph{Preserving behaviorally relevant color cues.}
We finally consider a context-dependent setting in which the correct action
depends on color information outside the randomized regions.
Figure~\ref{fig:kitchen_color_cue} shows a kitchen task in which the stove-light
color determines the required behavior: a green light instructs the robot to
remove the pot from the stove, whereas a red light instructs it to place the
plated butter into the pot. Full-image color randomization can alter this
decision-relevant cue during training and encourage the policy to ignore it.
MCR instead randomizes only the selected object regions, while RoboMP-DINOv2 retains the
unmodified scene context.

RoboMP-DINOv2-MCR achieves success rates of $88\%$ and $71\%$ in the green- and
red-light conditions, respectively, compared with $25\%$ and $19\%$ for
DP--DINOv2 trained with full-image color randomization. This experiment
demonstrates the behavior of the complete RoboMP-DINOv2-MCR pipeline in a setting that
requires robustness to task-object recoloring while preserving sensitivity to
a behaviorally relevant contextual cue. It also illustrates how RoboMP-DINOv2 provides
a suitable full-scene architecture for localized augmentation: MCR alters the
selected object regions without preventing the encoder from using the
unmodified stove-light signal.

\begin{figure}[t]
    \centering
    \setlength{\tabcolsep}{2pt}
    \begin{tabular}{ccc}
        \includegraphics[width=0.31\linewidth]{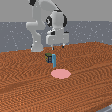} &
        \includegraphics[width=0.31\linewidth]{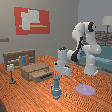} &
        \includegraphics[width=0.31\linewidth]{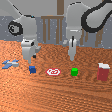} \\
        \includegraphics[width=0.31\linewidth]{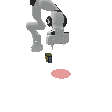} &
        \includegraphics[width=0.31\linewidth]{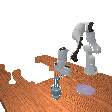} &
        \includegraphics[width=0.31\linewidth]{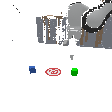}
    \end{tabular}
    \caption{Qualitative examples of imperfect evaluation-time mask predictions
    from Grounding DINO+SAM. Top row: RGB observations. Bottom row: predicted
    masks used by all mask-based methods. From left to right, the columns show
    Put Can (obstacle), Long Horizon, and Two-Arm stacking.}
    \label{fig:imperfect_masking}
\end{figure}

\clearpage
\bibliographystyle{plainnat}
\bibliography{references}

@inproceedings{huang2025roboground,
  title={Roboground: Robotic manipulation with grounded vision-language priors},
  author={Huang, Haifeng and Chen, Xinyi and Chen, Yilun and Li, Hao and Han, Xiaoshen and Wang, Zehan and Wang, Tai and Pang, Jiangmiao and Zhao, Zhou},
  booktitle={2025 IEEE/CVF Conference on Computer Vision and Pattern Recognition (CVPR)},
  pages={22540--22550},
  year={2025},
  organization={IEEE}
}

@inproceedings{emukpere2025disentangled,
  title={Disentangled Object-Centric Image Representation for Robotic Manipulation},
  author={Emukpere, David and Deffayet, Romain and Wu, Bingbing and Br{\'e}gier, Romain and Niemaz, Michael and Meunier, Jean-Luc and Proux, Denys and Renders, Jean-Michel and Kim, Seungsu},
  booktitle={2025 IEEE/RSJ International Conference on Intelligent Robots and Systems (IROS)},
  pages={15873--15879},
  year={2025},
  organization={IEEE}
}

@article{zhu2023learning,
  title={Learning generalizable manipulation policies with object-centric 3d representations},
  author={Zhu, Yifeng and Jiang, Zhenyu and Stone, Peter and Zhu, Yuke},
  journal={arXiv preprint arXiv:2310.14386},
  year={2023}
}

@article{hou2024diffusion,
  title={Diffusion transformer policy},
  author={Hou, Zhi and Zhang, Tianyi and Xiong, Yuwen and Pu, Hengjun and Zhao, Chengyang and Tong, Ronglei and Qiao, Yu and Dai, Jifeng and Chen, Yuntao},
  journal={arXiv preprint arXiv:2410.15959},
  year={2024}
}

@inproceedings{chi2023diffusionpolicy,
	title={Diffusion Policy: Visuomotor Policy Learning via Action Diffusion},
	author={Chi, Cheng and Feng, Siyuan and Du, Yilun and Xu, Zhenjia and Cousineau, Eric and Burchfiel, Benjamin and Song, Shuran},
	booktitle={Proceedings of Robotics: Science and Systems (RSS)},
	year={2023}
}

@misc{ren2024grounded,
      title={Grounded SAM: Assembling Open-World Models for Diverse Visual Tasks}, 
      author={Tianhe Ren and Shilong Liu and Ailing Zeng and Jing Lin and Kunchang Li and He Cao and Jiayu Chen and Xinyu Huang and Yukang Chen and Feng Yan and Zhaoyang Zeng and Hao Zhang and Feng Li and Jie Yang and Hongyang Li and Qing Jiang and Lei Zhang},
      year={2024},
      eprint={2401.14159},
      archivePrefix={arXiv},
      primaryClass={cs.CV}
}

@article{oquab2023dinov2,
  title={Dinov2: Learning robust visual features without supervision},
  author={Oquab, Maxime and Darcet, Timoth{\'e}e and Moutakanni, Th{\'e}o and Vo, Huy and Szafraniec, Marc and Khalidov, Vasil and Fernandez, Pierre and Haziza, Daniel and Massa, Francisco and El-Nouby, Alaaeldin and others},
  journal={arXiv preprint arXiv:2304.07193},
  year={2023}
}

@inproceedings{liu2023groundingdino,
  title={Grounding DINO: Marrying DINO with Grounded Pre-Training for Open-Set Object Detection},
  author={Liu, Shilong and Zeng, Zhaoyang and Ren, Tianhe and Li, Feng and Zhang, Hao and Yang, Jie and Li, Chao and Yang, Jianwei and Su, Hang and Zhu, Jun and Zhang, Lei},
  booktitle={European Conference on Computer Vision (ECCV)},
  year={2024}
}

@inproceedings{kirillov2023segment,
  title={Segment Anything},
  author={Kirillov, Alexander and Mintun, Eric and Ravi, Nikhila and Mao, Hanzi and Rolland, Chloe and Gustafson, Laura and Xiao, Tete and Whitehead, Spencer and Berg, Alexander C. and Lo, Wan-Yen and others},
  booktitle={Proceedings of the IEEE/CVF International Conference on Computer Vision (ICCV)},
  pages={4015--4026},
  year={2023}
}

@inproceedings{radford2021clip,
  title={Learning Transferable Visual Models From Natural Language Supervision},
  author={Radford, Alec and Kim, Jong Wook and Hallacy, Chris and Ramesh, Aditya and Goh, Gabriel and Agarwal, Sandhini and Sastry, Girish and Askell, Amanda and Mishkin, Pamela and Clark, Jack and Krueger, Gretchen and Sutskever, Ilya},
  booktitle={Proceedings of the 38th International Conference on Machine Learning (ICML)},
  pages={8748--8763},
  year={2021}
}

\end{document}